\documentclass[conference]{IEEEtran}

\usepackage{footnote}
\usepackage{amsmath}
\usepackage{graphicx}
\usepackage{tabularx}
\usepackage{url}
\usepackage{enumitem}

\begin{document}
%
\title{Fashion and Apparel Classification using Convolutional Neural Networks}

\author{
\IEEEauthorblockN{Alexander Schindler}
\IEEEauthorblockA{Austrian Institute of Technology \\
                       Digital Safety and Security \\
                       Vienna, Austria \\ alexander.schindler@ait.ac.at}
\and
\IEEEauthorblockN{Thomas Lidy}
\IEEEauthorblockA{Vienna University of Technology \\
               Institute of Software Technology \\
               Vienna, Austria \\ lidy@ifs.tuwien.ac.at}
\and
\IEEEauthorblockN{Stephan Karner, Matthias Hecker}
\IEEEauthorblockA{MonStyle \\
               Vienna, Austria \\ matthias.hecker@monstyle.io}
}


\maketitle

\begin{abstract}

We present an empirical study of applying deep Convolutional Neural Networks (CNN) to the task of fashion and apparel image classification to improve meta-data enrichment of e-commerce applications. Five different CNN architectures were analyzed using clean and pre-trained models. The models were evaluated in three different tasks \textit{person} detection, \textit{product} and \textit{gender} classification, on two small and large scale datasets. 

\end{abstract}

\section{Introduction}
\label{intro}

The recent progress in the image retrieval domain provides new possibilities for a vertical integration of research results into industrial or commercial applications. 
%
Based on the remarkable success of Deep Neural Networks (DNN) applied to image processing tasks, this study focuses on the task of fashion image classification. Photographs of clothes and apparels have to be classified into a set of pre-annotated categories such as skirt, jeans or sport-shoes. Online e-commerce companies such as Asos-EU \footnote{\url{http://www.asos.de/}}, Farfetch \footnote{\url{https://www.farfetch.com}} or Zalando \footnote{\url{https://www.zalando.de/}} provide access to the data of their products in stock including item-meta-data and images. 
%
Especially the provided meta-data varies in quality, granularity and taxonomy. Although, most of the companies provide categorical descriptions of their products, the applied terminology varies as well as the depth of the categorical hierarchy. Fashion image classification is thus used to consolidate the meta-data by enriching it with new generalized categorical labels.

%
This is a traditional image processing task with domain specific challenges of large variating styles, textures, shapes and colors. A major advantage is the image quality which are professionally produced high quality and high resolution images. There are generally two categories of photographs. The first arranges products in front of a white background. The second portraits a person or parts of a person who is wearing the products. While the first category reduces semantic noise of the images, the second one introduces it, because a person wearing multiple items such as jeans, t-shirt, shoes and belt is only assigned to a single label.
%
Clothing and apparel retrieval has been addressed to find clothes similar to a photograph \cite{liu2012street} or a given style \cite{di2013style}. The main challenge these studies faced was the definition and extraction of relevant features to describe the semantic content of the images with respect to the high variability and deformability of clothing items. Recent approaches harness the potential of Deep Neural Networks (DNN) to learn the image representation. In \cite{veit2015learning} a siamese network of pre-trained Convolutional Neural Networks (CNN) is used to train a distance function which can be used to asses similarities between fashion images.

%
In this study we present an empirical evaluation of various DNN architectures concerning their classification accuracy in different classification tasks. These tasks are evaluated on two different datasets on further two different scales. First, a wide evaluation is performed on a smaller scale dataset and the best performing models are then applied to large scale datasets.
%
The remainder of this paper is organized as follows. In Section \ref{rel_work} we review related work. In Section \ref{data} the datasets used for the evaluation are presented. Section \ref{dl_models} provides an overview of the evaluated neural network architectures. Section \ref{eval} describes the evaluation setup and summarizes as well as discusses the results. Finally, conclusions and outlooks to future work are given in Section \ref{conclusions}.

\section{Related Work}
\label{rel_work}

Recently, CBIR has experienced remarkable progress in the fields of image recognition by adopting methods from the area of deep learning using convolutional neural networks (CNNs). A full review of deep learning and convolutional neural networks is provided by \cite{chatfield2014return}. Neural networks and CNNs are not new technologies, but with early successes such as LeNet \cite{lecun1989optimal}, it is only recently that they have shown competitive results for tasks such as in the ILSVRC2012 image classification Challenge \cite{krizhevsky2012imagenet}. With this remarkable reduction in a previously stalling error-rate there has been an explosion of interest in CNNs. Many new architectures and approaches were presented such as \textit{GoogLeNet} \cite{szegedy2015going}, \textit{Deep Residual Networks (ResNets)} \cite{he2016deep} or the \textit{Inception Architecture} \cite{szegedy2015going}. Neural networks have also been applied to metrics learning \cite{jain2012metric} with applications in image similarity estimation and visual search.
Recently two datasets have been published. The MVC Dataset \cite{Liu2016} for view-invariant clothing retrieval (161.638) images and the DeepFashion Dataset \cite{liuLQWTcvpr16DeepFashion} with 800.000 annotated real life images.

\section{Data}
\label{data}

The data provided was retrieved from online e-commerce companies such as Asos-EU, Farfetch or Zalando. \\

\begin{figure}[t!]
\includegraphics[width=1.0\columnwidth]{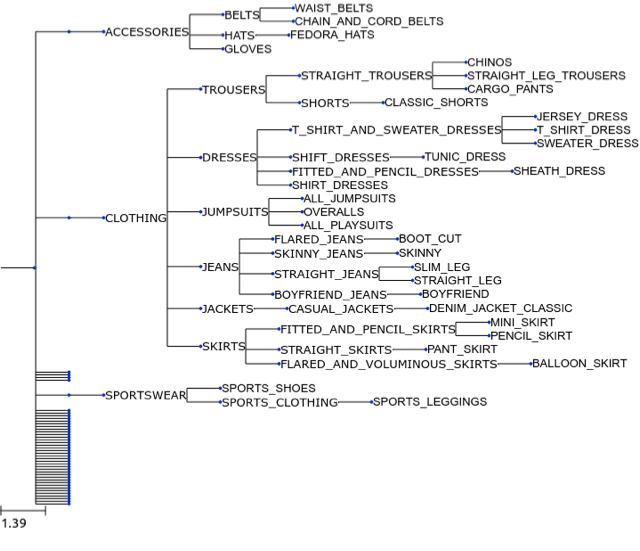}
\caption{Fashion categories hierarchy.}
\label{fig:class_hirarchy}
\end{figure} 

\noindent
\textbf{Person:} The persons dataset consists of 7833 images and the corresponding ground truth assignments. 5.669 images are labeled as \textit{Person} and 2.164 images are labeled as \textit{Products}. \\

\noindent
\textbf{Products:} The product dataset consists of 234.884 images and their corresponding ground-truth assignments. These images belong to 39.474 products where each product is described by 5,95 images on average. Ground-truth labels are provided for categories category, gender and age. All labels, including age, are provided on a categorical scale. The provided ground-truth assignments consists of 43 classes for the category attribute. These categories are based on a hirarchical taxonnomy. The hirarchy for the provided dataset is depicted in Figure \ref{fig:class_hirarchy}. Its largest class \textit{SPORTS SHOES} contains 66.439 images (10.807 products) and its smallest class \textit{JUMPSUITS} 6 images (1 product). To facilitate more rapid experimentation, the provided dataset was sub-sampled to approximately 10\% of its initial size. Further, due to the class imbalance of the provided category labels, an artificial threshold has been applied to the class sizes of the assignments. All classes with less than 100 images have been skipped. The remaining classes have been sub-sampled to a 10\% subset. The sub-sampling adhered to further restriction. First, stratification was used to ensure that the frequency distribution of class labels in the subsample corresponds to that of the original one. Second, sub-sampling was performed on product-level. This ensured the consistency of product-images and that there are no products with only one image. Finally, sub-sampling of a class was stopped when a minimum of 100 images was reached. This resulted in a subset of 23.305 instances, ranging from 5.659 images for \textit{SPORTS\_SHOES} (922 products) and 103 images for \textit{STRAIGHT\_LEG\_TROUSERS} (19 products). 


\section{Deep Neural Network Models}
\label{dl_models}

In this study we compared five different DNN architectures which varied in depth and number of trainable parameters, including three winning contributions to the ImageNet Large Scale Visual Recognition Challenge (ILSVRC) \cite{ILSVRC15} and two compact custom CNNs with fewer trainable parameters. 
The following architectures were evaluated:

\begin{itemize}[noitemsep, topsep=7pt, leftmargin=15pt]
\itemsep0.5em
\renewcommand\labelitemi{}
	
\item \textbf{Vgg16 and Vgg19:} very deep convolutional neural networks  (VGGnet) \cite{simonyan2014very} with 16/19 layers and 47/60 million trainable parameters, reaching an ILSVRC top-5 error rate of 6.8\%.
    
\item \textbf{InceptionV3:} high performance network at a relatively modest computational cost \cite{szegedy2015going} with 25 million trainable parameters reaching an ILSVRC top-5 error rate of 5.6\%.

\item \textbf{Custom CNN and Vgg-like:} compact convolutional neural network with only 10 million trainable parameters.
	
\end{itemize}

\begin{figure}[b!]
\includegraphics[width=1.0\columnwidth]{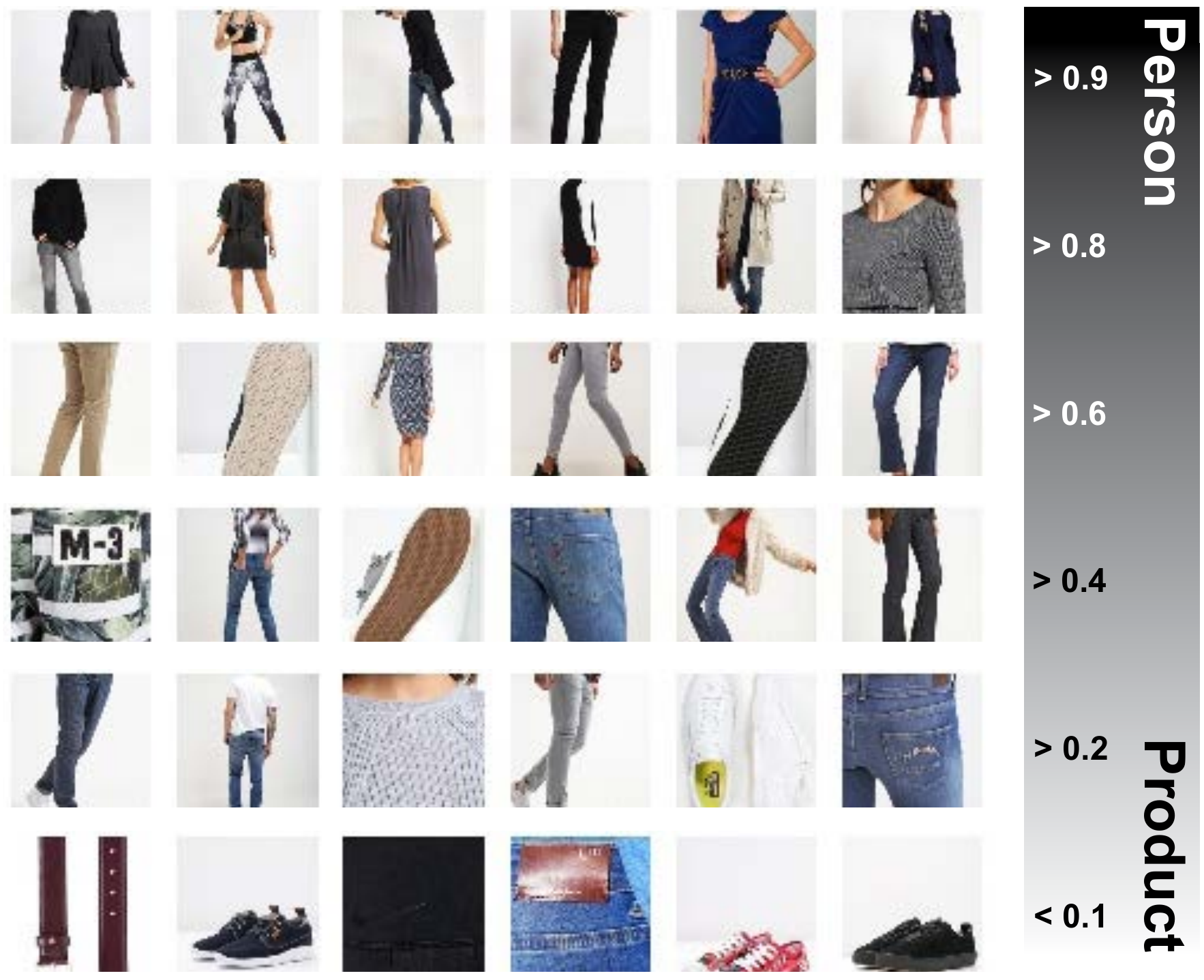}
\caption{Examples predictions of the \textit{person detector}. Prediction was realized as binary classification. Values above a values of 0.5 are classified as \textit{persons} and values below as \textit{products}. Images in the first line thus represent images predicted as \textit{persons} with high confidence.}
\label{fig:person_detect_examples}
\end{figure} 

\noindent
The models were implemented in Python 2.7 using the keras\footnote{\url{https://github.com/fchollet/keras}} Deep Learning library on top of the Theano\footnote{\url{https://github.com/Theano/Theano}} backend.

\section{Evaluation}
\label{eval}

The Convolutional Neural Networks were evaluated towards their classification accuracies in the tasks of differentiating \textit{persons from products} as well as classifying product images according their product \textit{category} and \textit{gender}. We performed three-fold cross-evaluation and calculated the accuracies on a per-image and a per-product scale. To calculate the per-product accuracy the cumulative maximum of all predicted product images was taken into account. 

\subsection{Detecting Persons}

Person detection was introduced based on the observation that products are presented in two general types. First, there are images of products placed in front of a white background or table. The other type of images are worn products. Because persons on these images are wearing more than one product such as trousers, shirts, shoes and belts, it is hard for a classifier to learn the right boundaries. Thus, the intention was to train a person detector and to either filter person images, or to use this additional information as input for further models.

We applied a custom VGG-like CNN with three layers of batch-normalized stacked convolution layers with 32, 64 and 64 3x3 filters and a 256 units fully connected layer with 0.5 dropout. We realized this task as a binary classification problem by using a sigmoid activation function for the output layer. Predictions greater or equal 0.5 were classified as persons. This approach already provided an accuracy of 91.07\% on the \textit{person} dataset.
Figure \ref{fig:person_detect_examples} shows example images of the person detection model. Images on the bottom row were predicted with values below 0.1 and are thus categorized as \textit{products}, whereas images in the top-row are considered to be \textit{persons}.

\subsection{Product Classification}

\begin{figure}[t!]
\includegraphics[width=1.0\columnwidth]{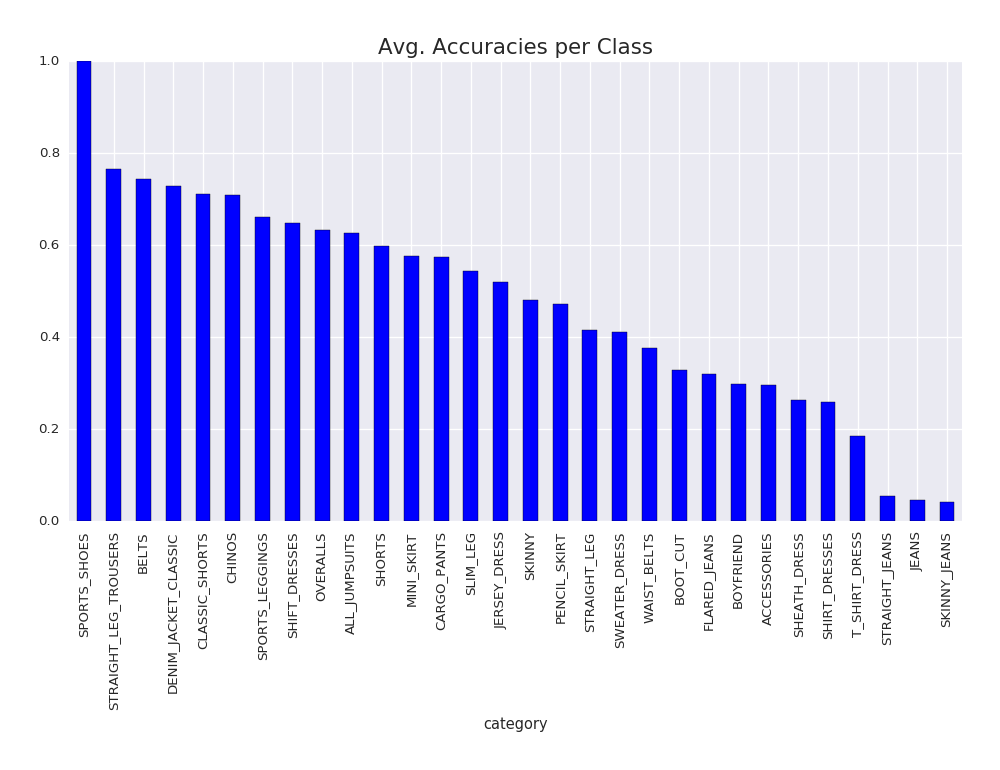}
\caption{Prediction accuracies on a per-image level for the best performing model - a fine-tuned \textit{InceptionV3} on the 234K dataset.}
\label{fig:class_accuracies}
\end{figure} 

The product classification experiments were conducted using the different CNN architectures presented in Section \ref{dl_models} on two different scales. First, a broad evaluation was performed on the small-scale subset of 23.305 images. Then, the best performing models were evaluated on the large scale 234.408 images dataset. All models, except those where explicitly mentioned, were trained using image data augmentation, including horizontal flipping of the image, shifting it by 25\% in height and width as well as a 25\% zoom range.

\subsubsection{Train from scratch or Fine-tune}

This part of the evaluation deals with the question of whether to train a model from scratch or to fine-tune a pre-trained model. The availability of a large collection of high quality images and a relative small number of classes suggests that models can be effectively fitted according the specific domain.

The results presented in Table \ref{tab:results_products} show that pre-trained models outperform clean models that have been specifically trained from scratch using only the images of the fashion image collection. Additionally, we evaluated the two different types of applying pre-trained models: a) resetting and training only the top fully connected layers while keeping all other parameters fixed, and b) continued fitting of all parameters on the new data - which is also referred to as fine-tuning. In either way the 1000 unit output layer of the pre-trained models had to be replaced with a 30 units layer representing the 30 product categories.The results of the evaluation show that fine-tuning outperforms the fitting of clean fully connected layers by 5.9\% (VGG16) to 7.9\% (InceptionV3).
The smaller custom models did provide an advantage concerning processing time of fitting and applying the model, but their accuracy results differ by 16.1\% to the top performing fine-tuned model.

\begin{figure}[b!]
\includegraphics[width=1.0\columnwidth]{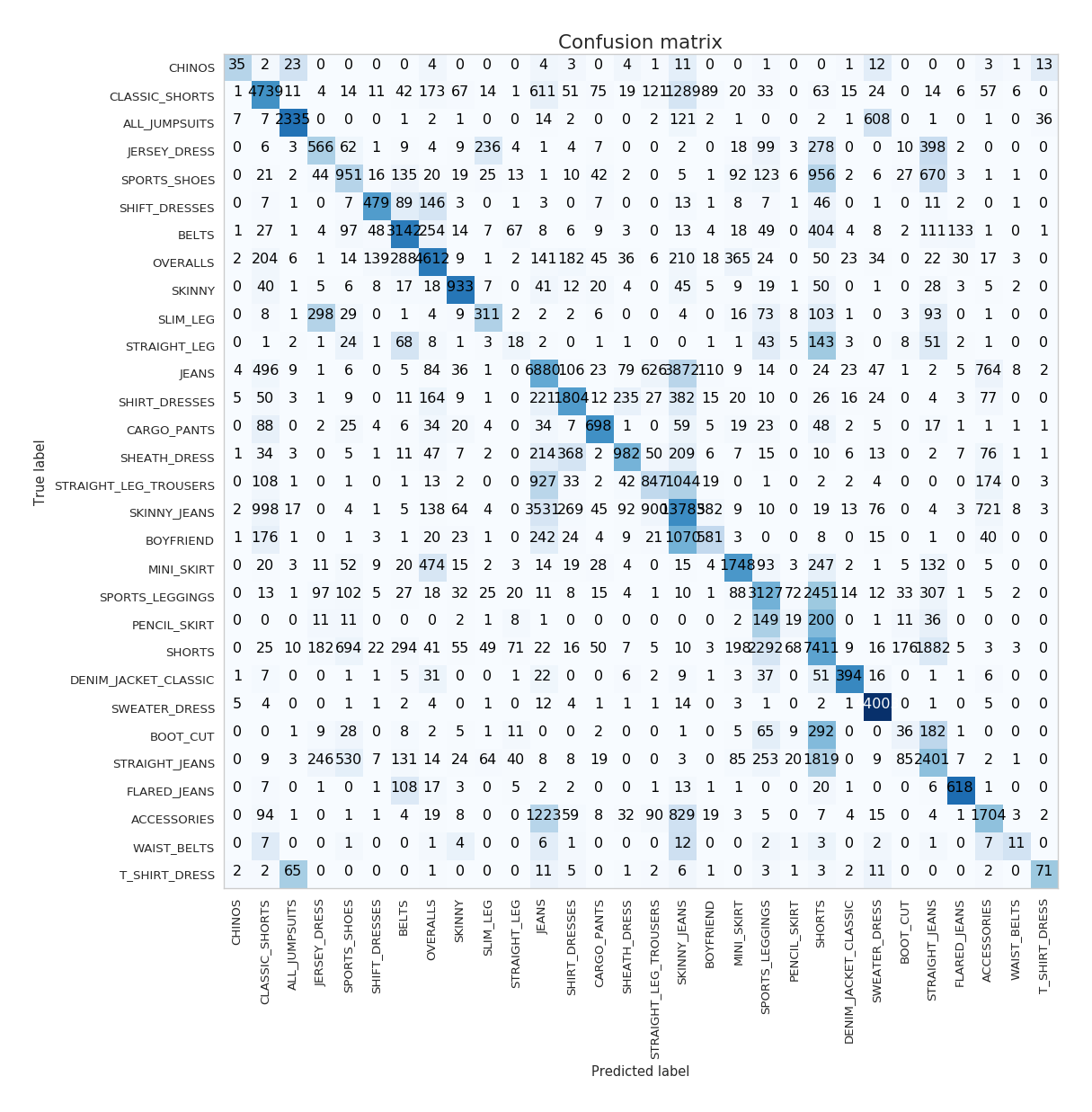}
\caption{Confusion matrix on a per-image level for the best performing model - a fine-tuned \textit{InceptionV3}  on the 234K dataset. The vertical axis represents the annotated category, the horizontal the prediction.}
\label{fig:confusion_matrix}
\end{figure} 

\begin{table*}[t!]
\centering
\begin{tabular}{ | l | l | l | l | }

	\hline
	\textbf{Description} & \textbf{best fold} & \textbf{best fold cum max} & \textbf{Mean cum max} \\ 
    \hline
    
    InceptionV3, pretrained, fine-tuned                               & 0.706 & 0.794 & 0.791 \\ 
    
    \hline
    
	InceptionV3, pretrained, fine-tuned                               & 0.658 & 0.729 & 0.716 \\ 
	VGG16, pretrained, fine-tuned                                     & 0.646 & 0.711 & 0.691 \\ 
	InceptionV3, pretrained, fine-tuned, person filter model as layer & 0.569 & 0.685 & 0.658 \\ 
	VGG19, pretrained, fine-tuned                                     & 0.579 & 0.673 & 0.634 \\
	InceptionV3, pretrained, fine-tuned, no augmentation              & 0.564 & 0.673 & 0.647 \\
	VGG19, pretrained, train only top-layers                          & 0.578 & 0.669 & 0.343 \\ 
	VGG16, pretrained, train only top-layers                          & 0.603 & 0.652 & 0.368 \\ 
	InceptionV3, pretrained, train only top-layers                    & 0.585 & 0.650 & 0.643 \\ 
	InceptionV3, pretrained, fine-tuned - person filtered metadata    & 0.640 & 0.636 & 0.614 \\ 
    InceptionV3, clean                                                & 0.492 & 0.594 & 0.580 \\
	Custom CNN, augmentation                                          & 0.506 & 0.568 & 0.538 \\
	Custom CNN                                                        & 0.463 & 0.556 & 0.523 \\ 
	Custom VGG-like                                                   & 0.438 & 0.549 & 0.519 \\ 
	VGG16, clean                                                      & 0.439 & 0.455 & 0.443 \\ 
	VGG19, clean                                                      & 0.437 & 0.447 & 0.430 \\ 
    
    \hline
    
    VGG19, pretrained, train only top-layers 	 &  0.819 &	0.887 &	0.880	\\
    InceptionV3, pretrained, fine-tuned &	0.798 &	0.863 &	0.836	\\
    VGG19, pretrained, fine-tuned   &	0.762 &	0.846 &	0.830	\\
    
	\hline

\end{tabular}
\caption{Classification results for the \textit{product category} classification task. Results summarize per image accuracy of the best fold, per product accuracy of the best fold, mean per product accuracy of all folds.}
\label{tab:results_products}
\end{table*}

Figure \ref{fig:class_accuracies} shows the prediction accuracy per class for the best model (fine-tuned IncepionV3) on the 234.408 images dataset. The most reliably predicted classes are \textit{SPORT\_SHOES}, \textit{STRAI\-GHT LEG TROUSERS} and \textit{BELTS}, the least reliable classes are \textit{STRAI\-GHT JEANS}, \textit{JEANS} and \textit{SKINNY\_JEANS}. These results indicate the problem of different granularity within the provided ground-truth assignments. Root- and leaf-nodes are used interchangeably which results from the aggregation of different e-commerce catalogs using different taxonomies. Although confusion a child- with a parent-class is semantically not wrong, but the trained models do not take this hierarchy into account and predict each label individually. Thiseffect can be seen in the confusion matrix in Figure \ref{fig:confusion_matrix} where spezialized classes such as \textit{JEANS} and \textit{SKINNY\_JEANS} or \textit{SKINNY} and \textit{SKINNY\_JEANS} or  are confused frequently.

\subsection{Gender Prediction}

The aim of the gender prediction task was to predict the intended gender of the product into the classes \textit{MALE}, \textit{FEMALE} and \textit{UNISEX}. The results are comparable to the product classification task in the sense that pre-trained and fine-tuned models provide the highest accuracies with a best performing value of 88\%.

\section{Conclusions and Future Work}
\label{conclusions}

In this study we presented an empirical evaluation of different Convolutional Neural Network (CNN) architectures concerning their performance in different tasks in the domain of fashion image classification.  The experiments indicated that dispite the large amount and high quality of provided fashion images, pre-trained and fine-tuned models outperform those which were trained on the given collections alone. Future work will concentrate on analyzing models on a scale of two million images.

\bibliographystyle{IEEEtran}

\bibliography{references}

\end{document}